# Assessment of electrical and infrastructure recovery in Puerto Rico following hurricane Maria using a multisource time series of satellite imagery

Jacob Shermeyer*[a]

[a]In-Q-Tel, CosmiQ Works, 2107 Wilson Blvd, Arlington, VA, USA 22003

## ABSTRACT

Puerto Rico suffered severe damage from the category 5 hurricane (Maria) in September 2017. Total monetary damages are estimated to be ~92 billion USD, the third most costly tropical cyclone in US history. The response to this damage has been tempered and slow moving, with recent estimates placing 45% of the population without power three months after the storm. Consequently, we developed a unique data-fusion mapping approach called the Urban Development Index (UDI) and new open source tool, Comet Time Series (CometTS), to analyze the recovery of electricity and infrastructure in Puerto Rico. Our approach incorporates a combination of time series visualizations and change detection mapping to create depictions of power or infrastructure loss. It also provides a unique independent assessment of areas that are still struggling to recover. For this workflow, our time series approach combines nighttime imagery from the Suomi National Polar-orbiting Partnership Visible Infrared Imaging Radiometer Suite (NPP VIIRS), multispectral imagery from two Landsat satellites, US Census data, and crowd-sourced building footprint labels. Based upon our approach we can identify and evaluate: 1) the recovery of electrical power compared to pre-storm levels, 2) the location of potentially damaged infrastructure that has yet to recover from the storm, and 3) the number of persons without power over time. As of May 31, 2018, declined levels of observed brightness across the island indicate that 13.9% +/- ~5.6% of persons still lack power and/or that 13.2% +/- ~5.3% of infrastructure has been lost. In comparison, the Puerto Rico Electric Power Authority states that less than 1% of their customers still are without power.

## INTRODUCTION

The tropical cyclone Hurricane Maria developed on September 16, 2017 in the Atlantic Ocean to the northeast of South America. The storm hit warm coastal waters and intensified rapidly, moving from a category 1 to a category 5 Hurricane according to the Saffir–Simpson hurricane wind scale (SSHWS) in less than 24 hours[1]. Maria subsequently moved in a northwesterly direction, making landfall in Dominica on September 19, and then eventually southeastern Puerto Rico, a territory of the United States, on September 20, 2017 with wind speeds of 155 miles per hour (Figure 1). After passing Puerto Rico, the storm moved north of Hispaniola, up the eastern coast of the mainland United States and then out to sea.

Damage to Puerto Rico was severe and widespread following the Hurricane, with heavy rainfall, flooding, storm surge, and high winds causing considerable damage. The power grid was mostly destroyed, causing a power outage for the entire population[2]. Electrical recovery for many locations has taken months, with estimates placing 45% of the population without power three full months after the storm[2]. Additionally, 95% of the cellular network, 85% of phone and Internet service was knocked offline[2]. Up to 50% of the population had no access to clean drinking water[2]. Total monetary damages are estimated to be ~92 billion USD, the third most costly tropical cyclone in US history since 1900[3]. Fatalities as a consequence of Maria are still under investigation, however the most recent estimates suggest between 793 to 8,498 excess deaths occurred following the storm[4].

Estimating the amount of damage through remote sensing techniques is cost effective and enables an alternative and independent assessment of the effects of the hurricane. Our method incorporates two kinds of remote sensing data: Suomi National Polar-orbiting Partnership Visible Infrared Imaging Radiometer Suite (NPP VIIRS) nighttime lights imagery as well as multispectral Landsat imagery. Several previous methods have used nighttime lights imagery to extract built up areas, assess power outages after storms, monitor economic output, and track population migrations. For example, Shi et al. (2014)[5], evaluated the effectiveness of NPP VIIRS and the Defense Meteorological Satellite Program's Operational Linescan System (DMSP/OLS) satellite to extract built-up areas in twelve cities in China via simple thresholding with an average overall accuracy of 89.58%. Cao et al. (2013)[6] used a NPP VIIRS nighttime lights time series to detect power outages after the June 2012 Washington DC derecho storm and October 2012 Hurricane Sandy event. This analysis was able to reliably identify areas and quantify the amount of significant power loss following each of these events for specific regions of interest. Li et al. (2013)[7], found a strong linear correlation ($R^2$ value of ~0.87 for all provinces) between VIIRS nighttime brightness and the economic index gross regional product in China. Gao et al. (2016)[8] quantified the amount of urban sprawl seen across China using a combination of DMSP and census data. Finally, Bharti et al. (2011)[9] quantify seasonal migrations and the linkages to the spread of measles in three cities in Niger using DMSP data.

*jshermeyer@iqt.org; phone 1 571 775-0328; cosmiqworks.org

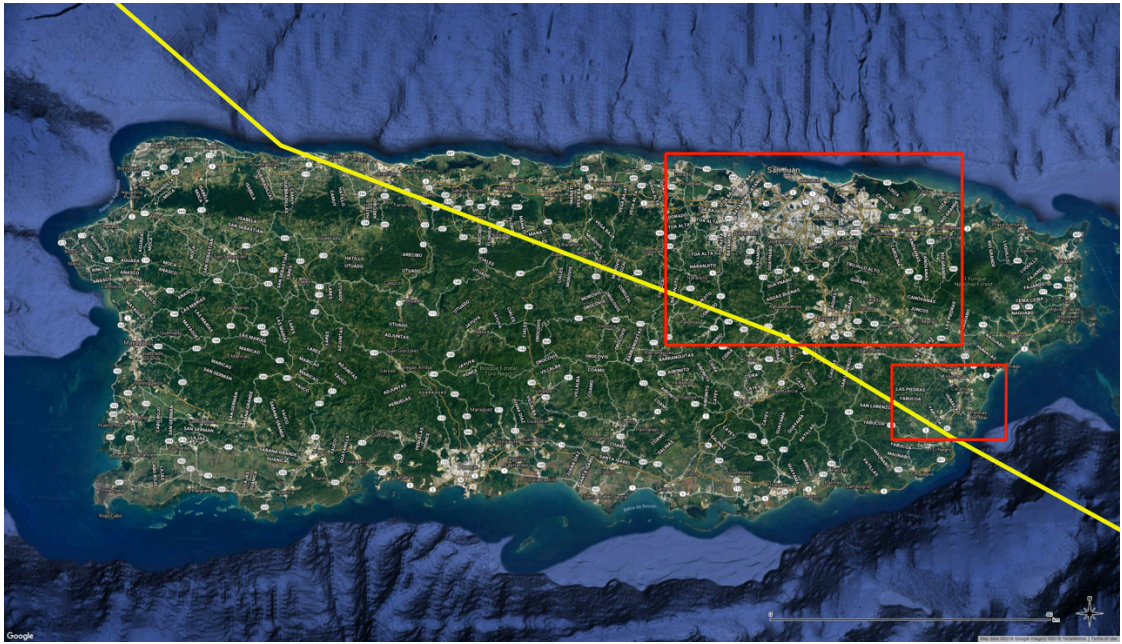

Figure 1. Map of the path of Hurricane Maria (yellow line), which moved from southeast to northwest and made landfall on September 20, 2017, and our two primary areas of interest that we investigate using time series visualizations in the results section (red boxes).

Landsat and multispectral imagery has been used to effectively map urbanization and impervious surface in hundreds of previous studies[10–23]. Typically, these studies follow a few direct paths to success, with most including some sort of machine learning classification approach, or the creation and application of an urban index from multiple spectral bands to derive urban land-cover. These methods are tried and true, and have been the backbone for the majority of modern land-cover remote sensing approaches. This paper takes a similar approach, and utilizes multispectral Landsat imagery to train a k-nearest neighbor (k-nn) classifier and derive new impervious surface maps. Although this machine learning approach is not a new technique, it does have unique aspects. Chiefly, this includes a transfer learning approach that leverages the most recent existing impervious surface map from 2001[10] to extract spectral signatures, and transfer them to create updated cloud free impervious surface maps from before Maria (January 2016-August 2017), and after Maria (September 2017-May 2018) for 98.9% of Puerto Rico.

Furthermore, this study combines these impervious surface maps with cloud and anomaly masked monthly composite NPP VIIRS nighttime lights imagery to remove misclassifications, enhance densely developed areas, and create time series visualizations at 30 meter resolution of urban development for the island. Additionally, VIIRS nighttime imagery alone was used to estimate brightness trends over time in all census tracts, and quantify the percentage difference in expected brightness from the observed brightness after the storm. These trends and differences are used to estimate the percentage of population without power, to estimate the amount of infrastructure lost, and to identify areas that are still struggling to recover. Although previous researchers have evaluated changes following storms, no studies have combined nightlights, multi-spectral imagery, census data, and building footprint labels to paint a holistic picture of the consequences of a storm. Overall, this research strives to address and evaluate the recovery of Puerto Rico over time from September 2017-May 2018.

# METHODS AND DATA

The methods and data are described throughout the following section. Generally, the methods fall under two categories: 1) We define a new quantity to measure impervious surface coverage, the Urban Development Index (UDI) and change map products (Figure 2), and 2) the evaluation of VIIRS nighttime brightness trends over time, and a comparison to US census tract population data[24] and Humanitarian OpenStreetMap (HOTOSM) building footprint labels[25] to estimate the number of persons without power or infrastructure loss.

## 1. Creation of the Urban Development Index

We created the Urban Development Index to evaluate and improve the quality of the maps that depict urban and suburban areas in Puerto Rico. The idea of the index is to combine an impervious surface map derived from Landsat imagery and monthly cloud free composites that depict nighttime brightness derived from NPP VIIRS data to validate one another and create a robust urban index. For example, the impervious surface maps that were created have false positives for impervious areas over highly reflective farmland. Conversely, VIIRS nighttime brightness mosaics suffer from light pollution in non-developed areas such as coastal waters and grassy areas. By multiplying these two images together, we can remove the false positives while quantifying the degree of development from densely urban to rural. Additionally, the finer spatial resolution of the impervious surface map helps to sharpen the VIIRS imagery to a 30m shared spatial resolution.

The impervious surface maps generated from the process described in section 2.2 depict impervious surface on a 1-10 scale with one being the least impervious (0-10%) and 10 being the greatest (90-100%) ($i$). VIIRS monthly cloud free composite data are described in section 2.3 and depict nighttime brightness radiance in Watts/cm$^2$. The units on these data range from no brightness (0) to very bright (>100) ($B$. Each VIIRS monthly cloud free composite image and the impervious surface map are multiplied together to produce the Urban Development Index that depicts urban development on a scale that roughly spans from 0-1000 (Equation 1). The UDI is calculated for each month from April 2017 to May 2018 to track changes over time and estimate true and percentage changes to the UDI.

$$UDI = i \, x \, B \qquad (1)$$

## 2. Landsat Data Description and Creation of Impervious Surface Maps

Two Worldwide Reference System 2 (WRS-2) path/rows (5/47 and 4/47) of Landsat imagery were identified to cover the majority (98.9%) of Puerto Rico and its' major metropolitan areas. All scenes were atmospherically corrected to surface reflectance using the Landsat Ecosystem Disturbance Adaptive Processing System (LEDAPS) or Landsat 8 Surface Reflectance (L8SR) system[26,27]. Atmospheric correction is required to standardize the images for the full time series and to ensure that using imagery from two different Landsat satellites does not compromise our results. All images were referenced in UTM coordinates and areas identified as medium to high confidence cloud or cloud shadow were masked out using the pixel QA layer. Each Landsat image in the time series includes only the Blue, Green, Red, Near Infrared (NIR), and both Shortwave Infrared (SWIR) bands.

Imagery was collected for two periods ranging from 2000-2002 and 2016-2018. The 2000-2002 time series included only Landsat 7 Enhanced Thematic Mapper+ (ETM+) imagery, as no Landsat 5 Thematic Mapper (TM) data were available. The 2016-2018 time series included only Landsat 8 Optical Land Image (OLI) data as the combination of cloud cover and scan line errors present in post-2003 Landsat 7 imagery available for Puerto Rico ultimately produced poor results. The 2000-2002 time series corresponds with the most recently created impervious surface map for Puerto Rico. This map depicts the percentage of impervious surface for the entire island for the year 2001[10].

The percentage of impervious surface is correlated to the amount of urban development and total population[10,16]. A transfer learning approach similar to the one described in Shermeyer and Haack (2015)[28] was devised to leverage the existing impervious surface map, and to create two updated impervious surface maps: pre-storm (January 2016 - August 2017), and post-storm from (September 2017 - May 2018).

First, the existing 0-100 percent impervious surface map from 2001 is segmented by increments of ten to generate ten distinct impervious surface classes. This simplifies the map and enables our classifier to more easily classify areas from little to highly impervious. This reclassified map is then used to extract spectral signatures from each of the cloud-masked Landsat images in the 2000-2002 time series. The average spectral response for every Landsat pixel that intersected with each of the ten impervious percentage classes is extracted and stored for each image. For each path/row an average spectral signature for each band is calculated as an aggregate of each image in the time series.

A k-nn classifier was then trained on these signatures and applied to each image in the pre-storm time series and the post-storm time series. The k-nn classifier was chosen primarily for its ease of implementation, for its exhibition of high accuracies in previous land cover mapping studies[28–30], and for its fast training and inference time (under one minute for each (64GB RAM CPU)). The value of k was set equal to one and the Euclidian distance metric with an equal neighbor-weighting scheme was chosen for this study. No fine-tuning of hyper-parameters was performed; instead relying on the strength of previous research that showed these parameters produced high accuracies for land-cover mapping[28–30]. It should be noted that tuning these parameters might have slightly increased the accuracy, precision, and recall of the final output maps. After each image in the pre and post-storm time series was classified into an impervious surface map, these maps were then aggregated together via averaging to create two cloud-free composite pre and post-storm impervious maps. Each map serves as an input into the pre and post storm urban development index change product.

### 3. NPP VIIRS Data Description

Monthly composites of NPP VIIRS day/night band (DNB) imagery were chosen as the primary nighttime imagery input for this workflow. VIIRS monthly composites are cloud masked and filtered to exclude areas that are affected by stray light, lightning, and lunar illumination[31,32]. The masked daily images are then averaged together for each month to create these composite products and have a nominal spatial resolution of ~450 m over Puerto Rico when using the UTM Zone 20N projection. These data presently span from April 2012 - May 2018, and any analysis was conducted over this period or portions of it. The data are distributed with two images, the first depicts nighttime brightness in floating point radiance values with units in Watts/cm$^2$, and the second depicts the number of observations acquired for this month. Areas with zero observations for a month due to cloud cover or other anomalies are masked out. The full 2012 - 2018 time series of imagery was used for evaluations in change of brightness over time and to evaluate changes to brightness following hurricane Maria. Individual monthly images for the months following Maria (September 2017 - April 2018) were used to create new UDI maps for each month to evaluate the location and speed of recovery.

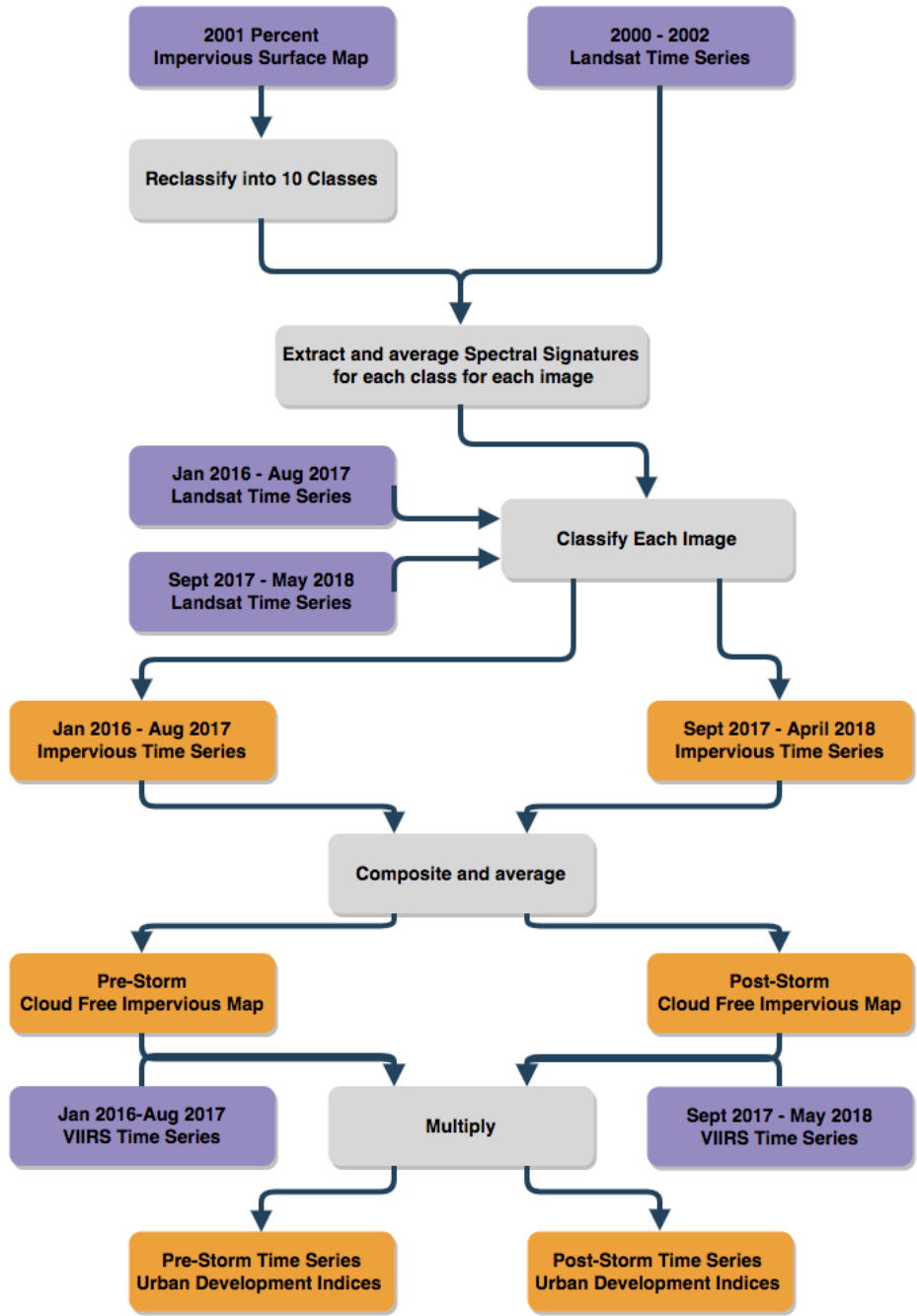

Figure 2. Flow chart of the Urban Development Index workflow. Inputs are colored in purple, actions in gray, and outputs in orange.

4. **Electrical and Infrastructure Recovery Analysis**

Comet Time Series (CometTS) was developed to facilitate analysis, extract relevant statistics, and enable visualization for a time series of satellite imagery in any user defined area of interest[33]. This tool has previously been used to estimate changes in population based upon changes in brightness to nighttime lights in Africa and Europe[34]. For this study, the tool was used to extract the change in average nighttime brightness for all census tracts within Puerto Rico and to infer electrical recovery and infrastructure loss. The tool uses a zonal statistics approach to extract various statistics on brightness within each census tract.

An autoregressive integrated moving average (ARIMA)[35] time series trend analysis was then conducted on the output of the tool for each census tract. A 5-month centered moving average (CMA) is calculated for each month in the time series from April 2012-August 2017. Irregularities from the smoothed CMA and the average monthly trends are calculated. These components are combined with the smooth CMA and a linear regression to create a seasonal forecast trend line for September 2017-May 2018. This trend line is used to determine where brightness is expected to be if the hurricane had not hit. The difference between the actual observed brightness and the forecast brightness are used to quantify electrical and infrastructure deficiencies and recovery. The mean absolute deviation or error from the seasonal trend line is also calculated to quantify what the expected range in error should be and to determine if the difference between the actual brightness and the projected brightness is significant. Finally this trend analysis is compared against 2016 census population estimates[24] to approximate the number of persons without power and against the HOTOSM building footprint[25] counts by census tract to approximate the percentage of buildings lost.

# RESULTS AND DISCUSSION

5. **Urban Development Index Visualizations**

An analysis of the urban development index depicts a precipitous drop off in UDI following Hurricane Maria, which made landfall in Puerto Rico on September 20 2017. A 264 point stratified random sample was created and visually interpreted with high-resolution imagery to test the accuracy of the UDI. Results indicate the index can effectively identify urban development with an accuracy of 90.9%. Unsurprisingly, smaller rural buildings and less densely developed areas are harder to map than highly developed areas. The San Juan metropolitan area and Caguas were chosen for this visualization as they represent the highest density urban development, with ~46% of the population living in these areas[24]. Additionally, these cities are quite representative of the rest of the island, which experienced similar damage, particularly in the central and eastern regions. Changes in UDI for San Juan and Caguas are depicted in videos 1 and 2 in this section. Video 1 depicts the effects of this storm and the changes to the UDI over time; video 2 depicts the differencing of a March-August 2017 average composite UDI from September 2017- May 2018 monthly UDIs. As the storm hit late in the month of September, only a small change can initially be seen in the September visualization. However, in October, the largest drop occurs. During this month, national estimates place 100% of the population without power[2]. Power and infrastructure gradually begin to recover, but are still not fully recovered by May 2018. Pocketed communities across the city have a much lower UDI, indicating a loss of power or infrastructure. A few select areas are brighter, mostly including the airport, ports, and a few select parks and open green spaces.

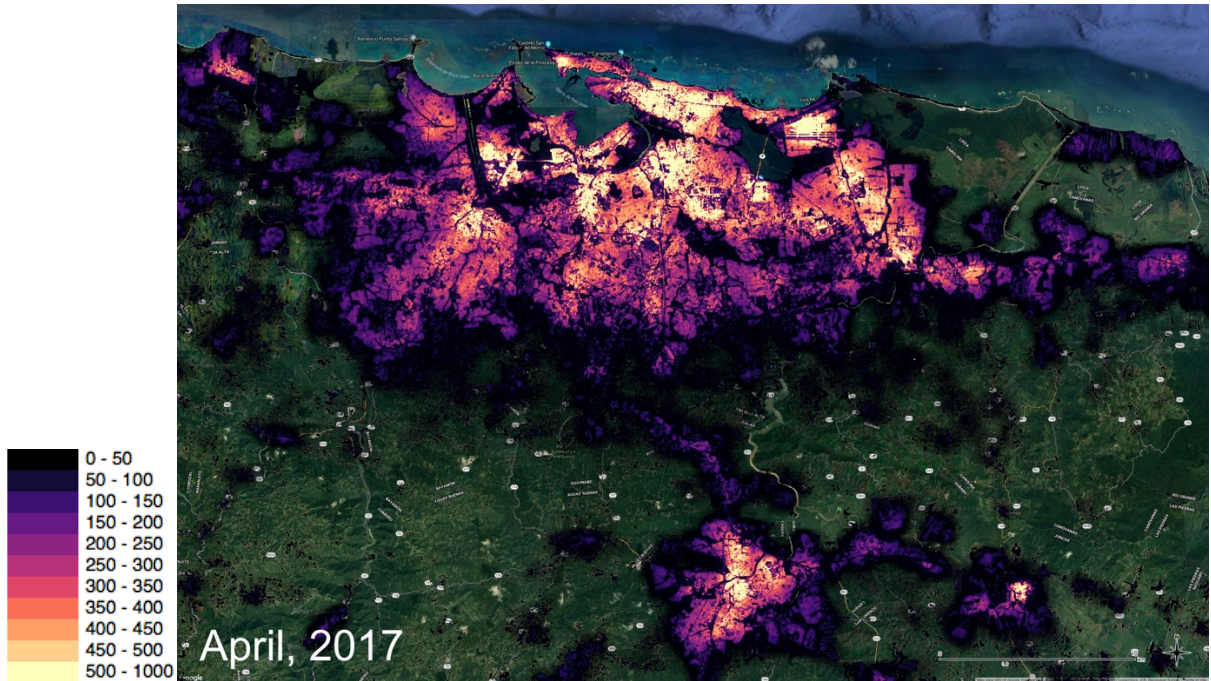

Video 1. This video depicts a visualization of change in the Urban Development Index from April 2017 - May 2018 for San Juan and Caguas, Puerto Rico. Areas that have a UDI value of zero are made transparent. All visualizations are overlaid on high-resolution imagery supplied via Google. Play Video

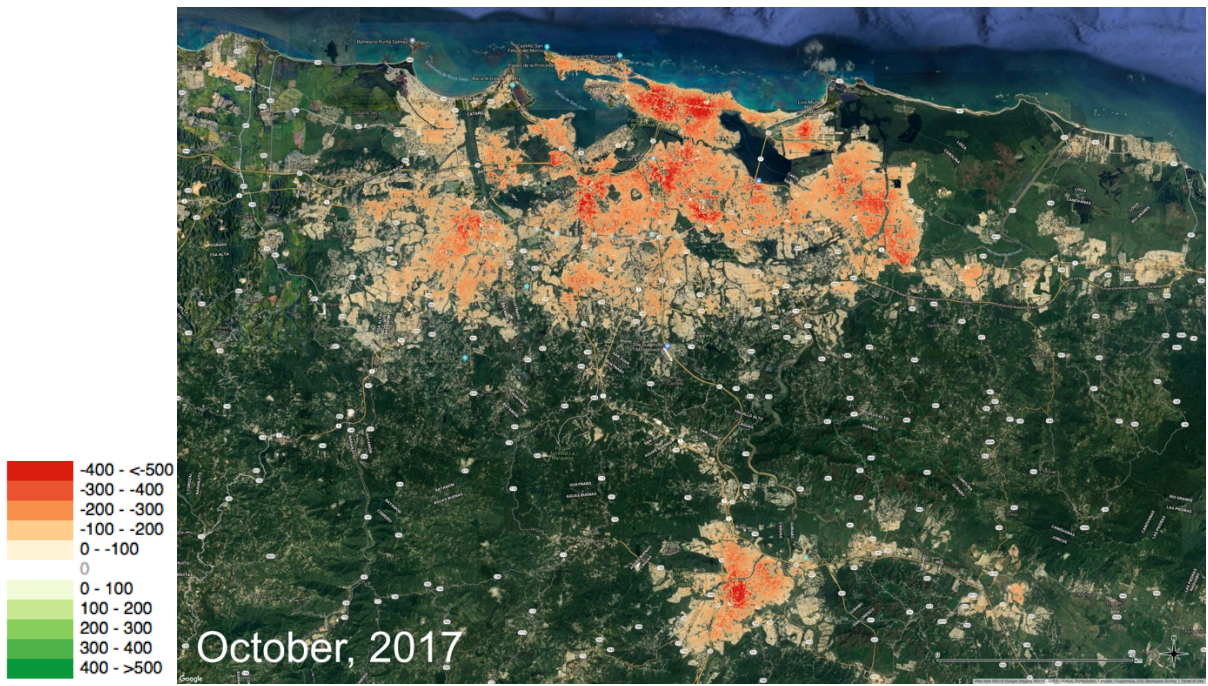

Video 2. This video depicts a visualization of changes over time for via subtracting a pre-storm composite UDI from monthly UDIs for San Juan and Caguas, Puerto Rico. Areas that have a UDI change value of zero are made transparent. All visualizations are overlaid on high-resolution imagery supplied via Google. Play Video

The southeastern portion of the island, the city of Humacao, and surrounding rural communities were also chosen for investigation, as this was one of the most severely affected regions. Changes in UDI are depicted in videos 3 and 4 in this section. Video 3 depicts the effects of this storm and the changes to UDI over time; video 4 depicts the percentage change from March-August 2017 versus the current month. Again, changes in October were the greatest with power and infrastructure gradually recovering thereafter.  As of May 2018, the rural communities in the western-most forested region still do not have power, and/or the buildings here have been destroyed.  Most of the region remains darker as well with the exception of coastal resorts in the southeast, the industrial park to the northwest, and the university in the western portion of Humacao.  These areas are likely the first to repaired, and ultimately recover, with some nighttime construction increasing brightness.

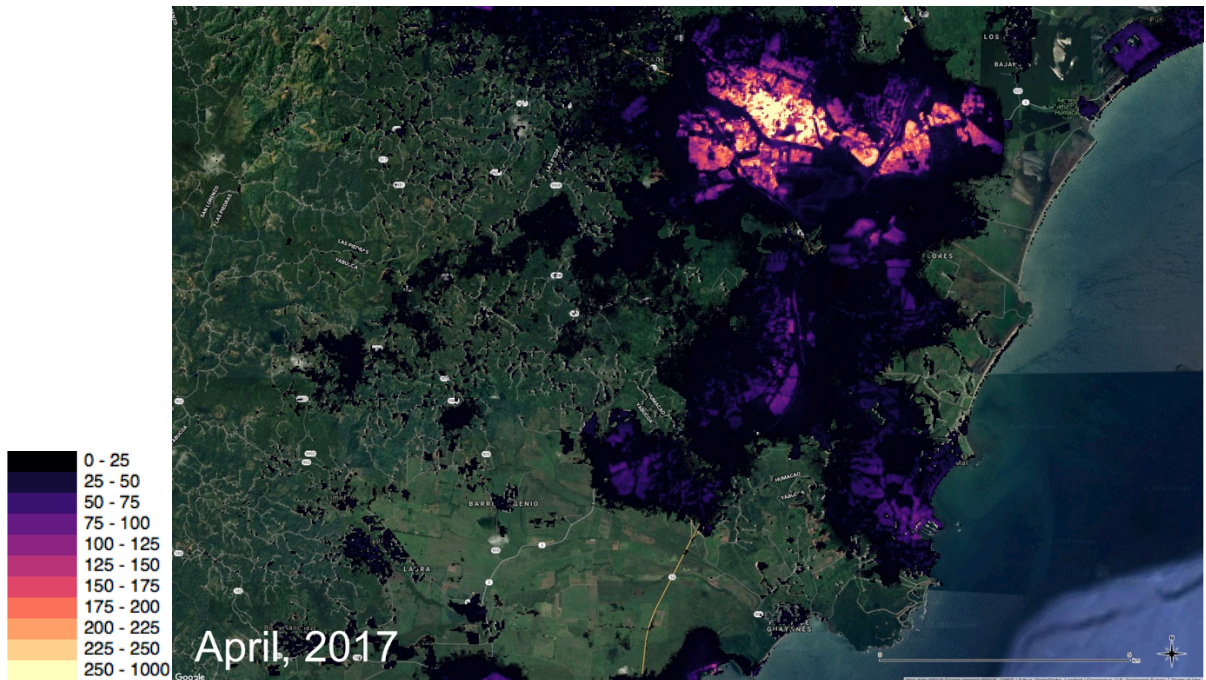

Video 3.  This video depicts a visualization of change in the Urban Development Index from April 2017 - May2018 for Humacao, Puerto Rico and the surrounding rural communities.   Areas that have a UDI value of zero are made transparent.  All visualizations are overlaid on high-resolution imagery supplied via Google.  Play Video

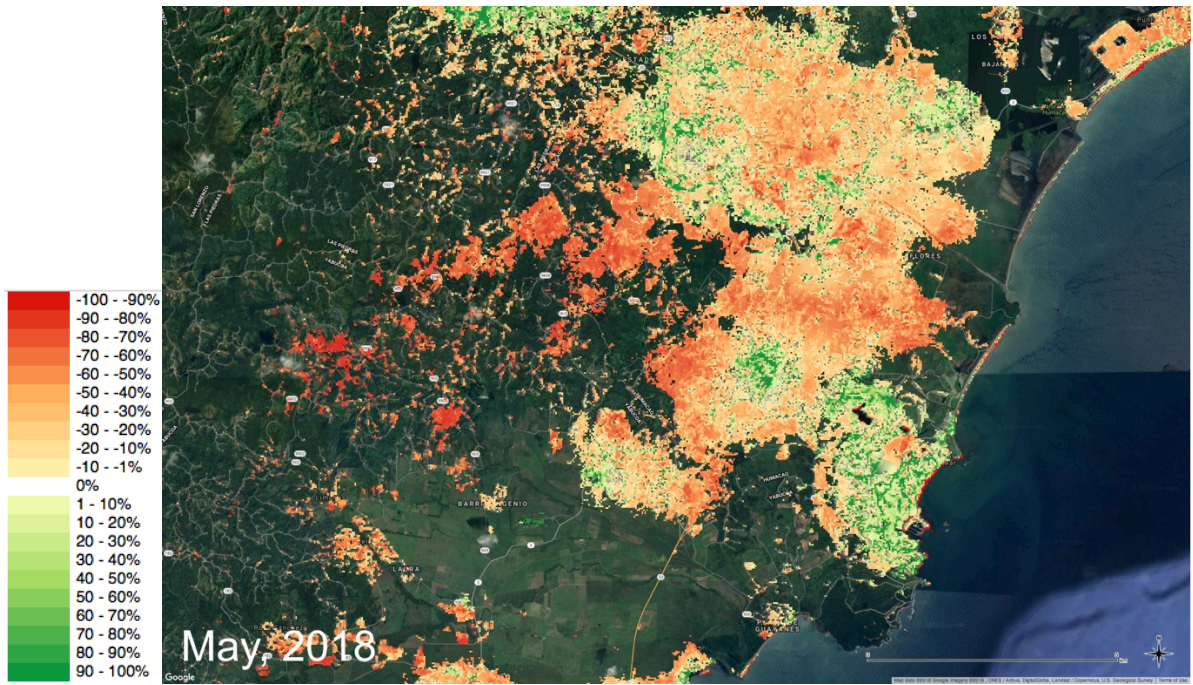

Video 4. This video depicts a visualization of the percent difference of monthly UDIs versus a pre-storm composite UDI for Humacao, Puerto Rico and the surrounding rural communities. Areas that have a UDI change value of zero are made transparent. All visualizations are overlaid on high-resolution imagery supplied via Google. Play Video

6. **Full Time Series Evaluation by Census Tract**

The percentage difference between projected VIIRS brightness and observed VIIRS brightness for each census tract are shown for all of Puerto Rico from April 2017 to May 2018 (Video 5). This analysis shows that observed versus expected brightness levels were generally in agreement before the storm, with some tracts being slightly brighter than expected (April-August 2017). A mild drop off brightness follows in September 2017, followed by a rapid drop off in October 2017. The mild September drop off is due to the nature of the monthly composite VIIRS imagery. Only a few observations are available after the storm hits (September 20th), and are equally weighted against pre-storm observations. Consequently, in October we begin to see the full magnitude of the impacts of Maria with the majority of the island, particularly in the east, showing a decline in observed versus projected brightness of 50% or greater. Power restoration begins to show in November and December 2017 with parts of San Juan coming back online, and the western coastal census tracts recovering first. More stability can be witnessed returning in January and February of 2018, with the exception of the southeast and interior rural areas of the island. Finally, through March, April, and May we can see a return to normalcy for most of the western portion of the island. The eastern portion still falls short of the forecast brightness with the majority of the tracts falling between 10 to 30% short of projection.

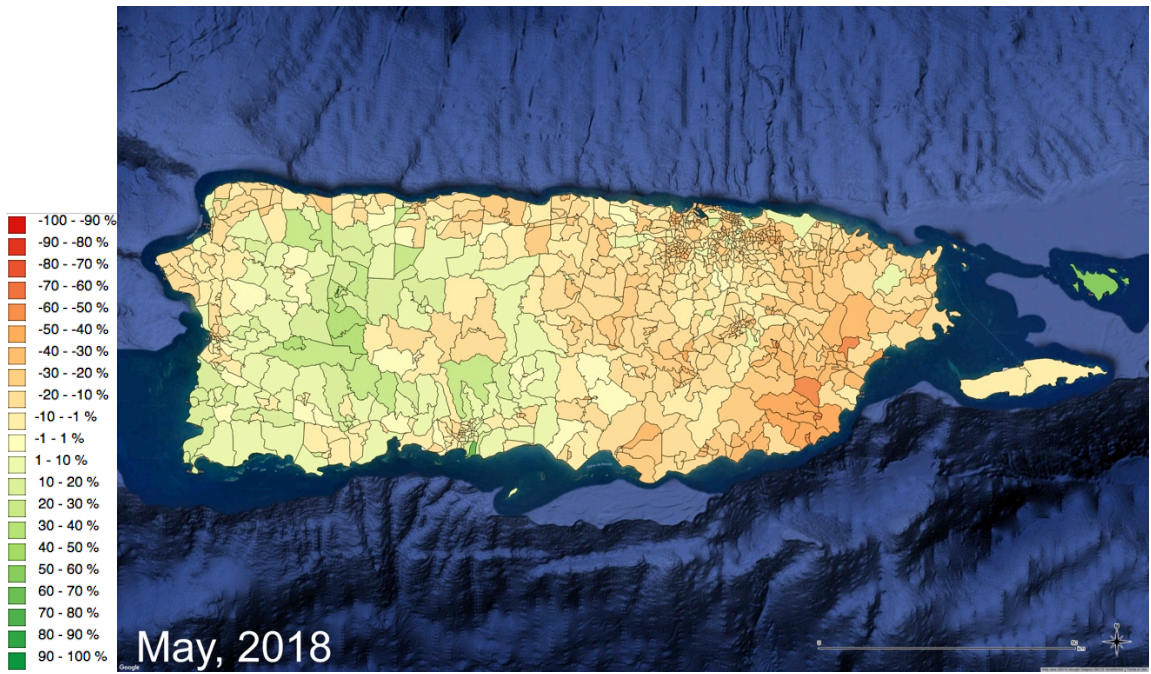

Video 5. This video depicts a visualization of the percentage difference between observed brightness and the ARIMA derived seasonal trend line and forecast from April 2017 – May 2018 for all census tracts in Puerto Rico. All visualizations are overlaid on high-resolution imagery supplied via Google.  Play Video

Time series visualizations were also created for all tracts.  Figures 3 and 4 depict two tracts that suffered severe damage, and have fallen far short of recovery.  These figures show what the map (Video 5) is visualizing in plot format.  The black points are the mean observed VIIRS brightness levels, the gray bounding area is one standard deviation of brightness within that tract, the orange trend line is the ARIMA based forecast that depicts the expected brightness if Maria had not hit, and finally the teal line represents a liner regression forecast.  Tract 9509 in Yabucoa is located in the southeastern portion of the island, it is one of the first locations the storm made landfall.  Its' recovery is also visualized in videos 3 and 4.  Tract 316.12 is in the Bayamon Municipio, located in the western portion of San Juan and seen in videos 1 and 2.  Both of these visualizations show tracts that still have only marginally recovered to expected brightness levels.  This is a result of either a lack of power restoration, the loss of infrastructure, and/or the loss of population.

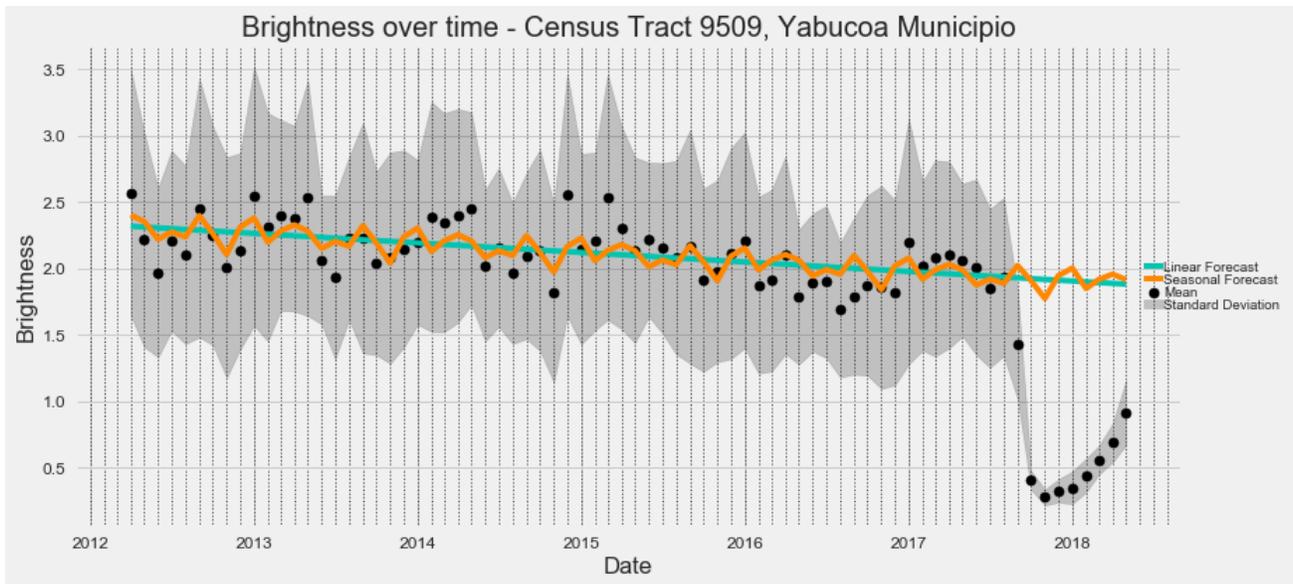

Figure 3. A visualization of mean and one standard deviation of brightness over time in Census Tract 9509, in Yabucoa Municipio. The linear regression forecast and seasonal adjusted forecast are plotted in teal and orange respectively.

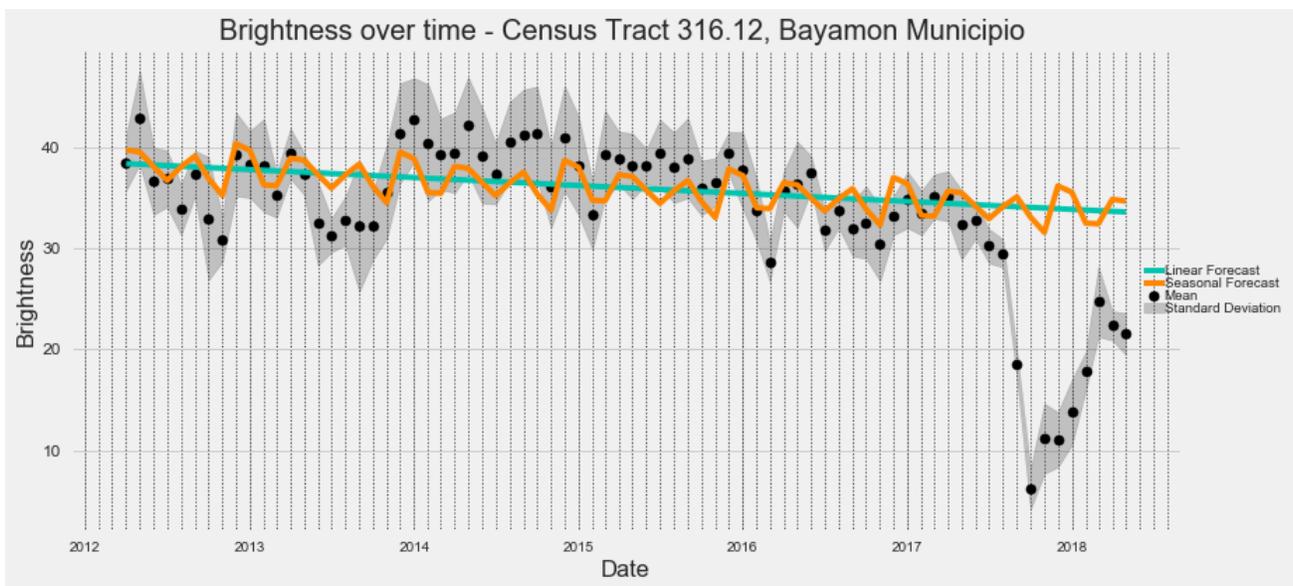

Figure 4. A visualization of mean and one standard deviation of brightness over time in Census Tract 316.12, in Bayamon Municipio. The linear regression forecast and seasonal adjusted forecast are plotted in teal and orange respectively.

## 7. Estimates of Electrical and Infrastructures Loss

An estimate of the number of persons that could potentially still lack power was also conducted (Figure 5). By estimating the percentage short of the ARIMA seasonal forecast trend line and multiplying that value by the census tract population estimate[24], we can quantify how many persons likely were or still are without power. In this figure, the VIIRS based estimate of persons without power is plotted in black with the mean absolute deviation from our ARIMA trend analysis from before the storm visualized in gray. The red line is the PREPA estimate for the number of customers without power[2,36]. Customers are likely not an absolute identifier for the number of persons without power, but they are correlated.

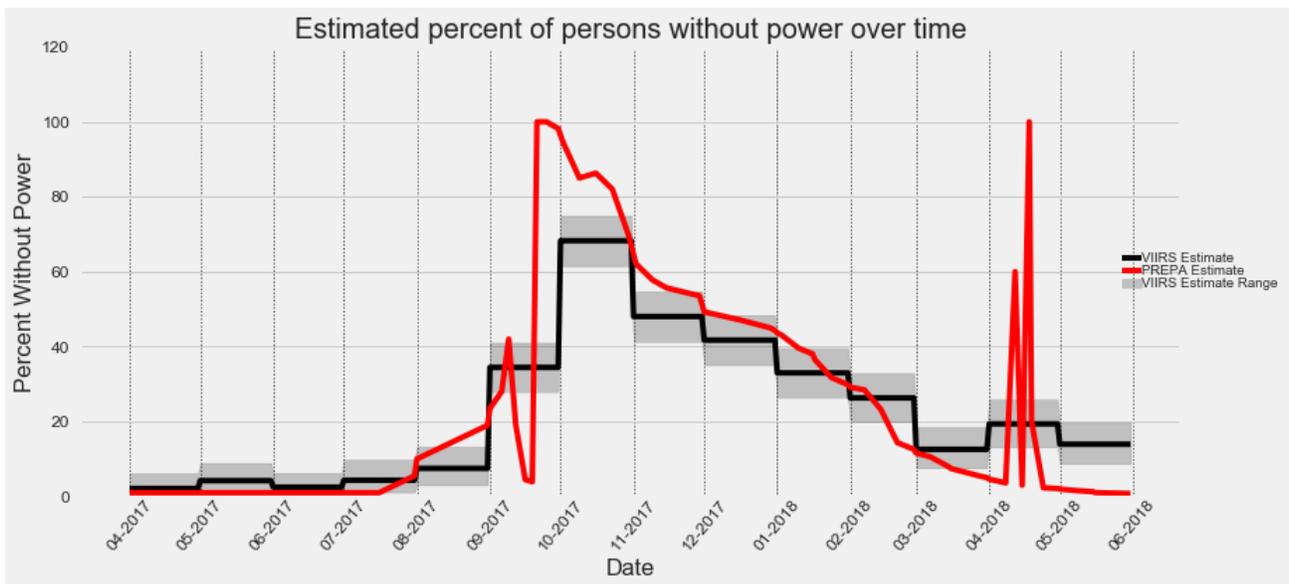

Figure 5. A visualization of the estimated persons without power across all of Puerto Rico. PREPA estimates[2,36] are in red, and VIIRS derived estimates are in black. Any gaps between PREPA estimates are interpolated. The mean absolute deviation from the ARIMA derived seasonal trend line before the storm is visualized in gray.

The main conclusion is that there is generally strong agreement between the VIIRS based estimates and PREPA estimates with a few clear exceptions, particularly from April to May 2018. The VIIRS estimates for the number of persons without power are slightly lower than the PREPA estimates for the beginning of October, again, likely because of the lack of cloud free observations during this period, artificially increased brightness because of nighttime repair activities, or an overestimation by PREPA of the number of persons without power. PREPA estimates and the VIIRS derived estimates then track closely to one another through November 2017 to mid-March 2018. In April, two sudden massive outages occurred, the first from a falling tree and the second due to a bulldozer backing into a power line[37]. Repairs were made quickly and power was restored within roughly 24 hours following these events. However, this highlights the fragility of the current grid, with two minor events causing nearly the entire island to lose power. As this workflow used VIIRS monthly composites, these single day outages are weighted against the average brightness for the rest of the month, meaning only a small increase in estimated the percentage of persons without power for April versus March and May of 2018. Ultimately, estimates between the VIIRS analysis and PREPA begin to diverge in April and May of 2018. By the end of May 2018, the PREPA estimates the total number without power to be less than 1%[36]. However, the VIIRS based analysis estimates that 13.9% +/- ~5.6% remain without power on the island. A comparative analysis versus HOTOSM building footprint counts by census tract indicates that 13.2% +/- ~5.3% of infrastructure may have been lost.

There are several conclusions that can be drawn from this analysis. The first is that the PREPA appears to be underestimating the number of persons that still do not have access to electricity. Ultimately, it is likely difficult for the PREPA to quantify the number of customers that have power, particularly in the rural areas in the southeast and central parts of Puerto Rico. The number of customers without power and the number of persons without power are similar but are not exact correlates. Although power grids may be 99% repaired, a large amount of destroyed infrastructure is still not receiving power, certainly leaving many persons still in the dark. The difference between the population and building analyses (0.69%) closely aligns with PREPA estimates of those without power (0.78%). Regardless, if the infrastructure has been destroyed, it is not receiving power; we can conclude that approximately 13% of residents still likely lack electricity. Another conclusion is that the decline in brightness is caused by a population decline and large emigration of persons out of Puerto Rico. The emigration of the population out of Puerto Rico has been shown to be aggressive. Melendez and Hinojosa (2017)[38] estimate that up to 14% of the population will leave by December 2019, primarily as a result of Maria. The government of Puerto Rico estimates that 200,000 will have left by the end of 2018, approximately 6% of the population. As of May 2018, this likely means a loss of 2-5% of the population. This mass emigration certainly will cause some level of decline in power consumption, and brightness as a corollary. Overall, the most likely outcome is a combination of these conclusions; however, we can safely state that the observed decline in brightness does not agree with PREPA estimates of the amount of persons or buildings receiving and/or using power.

## CONCLUSION

The aftermath of Hurricane Maria is still being felt across Puerto Rico at large. Our independent assessment relied on a fusion of remote sensing data, population estimates, and building footprint labels to quantify the number of persons without electricity and the amount of infrastructure lost. This research leveraged a new tool called CometTS to facilitate analysis, extract relevant statistics in census tracts, and enable visualization of a time series of NPP VIIRS nighttime lights imagery. Our findings showed that as of May 31, 2018, declines in brightness levels across the island indicate that 13.9% +/- ~5.6% of persons still lack power and/or that 13.2% +/- ~5.3% of infrastructure has been lost. The PREPA states that less than 1% of their customers still are without power, which appears to be a substantial underestimation based upon our findings. Other factors such as a mass emigration of persons out of Puerto Rico and an ongoing economic downturn are also contributing to the decline in brightness across the island. Comprehensively, time series visualizations and maps show several areas that are not receiving electricity and a large amount of unrepaired infrastructure damage in both urban and rural communities.